\documentclass[sigconf,authorversion,nonacm]{acmart}
\AtBeginDocument{%
  \providecommand\BibTeX{{%
    \normalfont B\kern-0.5em{\scshape i\kern-0.25em b}\kern-0.8em\TeX}}}

\usepackage{xspace}
\usepackage{xcolor}
\usepackage{color, colortbl}
\usepackage{caption}
\usepackage{subcaption}

\definecolor{LightCyan}{rgb}{0.88,1,1}
\usepackage[]{algorithm2e}
\usepackage{stfloats}
\definecolor{mygreen}{rgb}{0,0.6,0}

\newcommand{\eg}{e.g.,}

\newcommand{\ex}[1]{``\emph{#1}''}

\newcommand{\User}{\textcolor{orange}{User}}
\newcommand{\xTalkBot}{\textcolor{blue}{Bot}}
\newcommand{\PosDiff}[1]{\textcolor{mygreen}{#1}}
\newcommand{\NegDiff}[1]{\textcolor{red}{#1}}

\newcommand{\calA}{\mathcal{A}}

\newcommand{\calX}{\mathcal{X}}

\setcopyright{none}
\begin{document}

%% Consider defining shortitle as well.
\title{Dynamic Planning in Open-Ended Dialogue using Reinforcement Learning}
\author{Deborah Cohen, Moonkyung Ryu, Yinlam Chow, Orgad Keller, Ido Greenberg, Avinatan Hassidim, Michael Fink, Yossi Matias, Idan Szpektor, Craig Boutilier, Gal Elidan}
\email{{debbycohen,mkryu,yinlamchow,orgad,idogreenberg,avinatan,fink,yossi, szpektor,cboutilier,elidan}@google.com}
\affiliation{                               
\institution{Google Research}
}

% Concise list of authors for headers.
\renewcommand{\shortauthors}{Cohen et al.}

\begin{abstract}
Despite recent advances in natural language understanding and generation, and decades of research on the development of conversational bots, building automated agents that can carry on rich open-ended conversations with humans ``in the wild'' remains a formidable challenge. In this work we develop a real-time, open-ended dialogue system that uses reinforcement learning (RL) to power a bot's conversational skill at scale. Our work pairs the succinct embedding of the conversation state generated using SOTA (supervised) language models with RL techniques that are particularly suited to a dynamic action space that changes as the conversation progresses. Trained using crowd-sourced data, our novel system is able to substantially exceeds the (strong) baseline supervised model with respect to several metrics of interest in a live experiment with real users of the Google Assistant.
\end{abstract}

%% Please copy and paste the code instead of the example below. CCSXML
% \copyrightyear{2020}
% \acmYear{2020}
% \setcopyright{iw3c2w3g}
% \acmConference[WWW '20]{Proceedings of The Web Conference 2020}{April 20--24, 2020}{Taipei, Taiwan}
% \acmBooktitle{Proceedings of The Web Conference 2020 (WWW '20), April 20--24, 2020, Taipei, Taiwan}
% \acmPrice{}
% \acmDOI{10.1145/3366423.3380167}
% \acmISBN{978-1-4503-7023-3/20/04}


\maketitle

\section{Introduction}

With tremendous advances in AI and ML techniques to recognize speech and perform high quality natural language understanding (NLU) and generation (NLG), increased attention is being directed toward the task of carrying out real-time, rich conversations between humans and bots (e.g., \cite{li2016deep,Serban2017Deep,zhou2020design}). Realistic interactions generally span complex topic spaces and are relatively open-ended, and often have an underlying goal (e.g., task completion, knowledge sharing). Thus, carrying them out effectively requires not just powerful bots that learn to generate favorable responses, but also demands that bots have the ability to plan and adapt on the fly.

The framework of \emph{reinforcement learning (RL)} is a natural approach for this task. Indeed, research on \emph{Markov decision processes (MDPs)} and RL for spoken-dialogue systems spans well over two decades, ranging from early work using MDPs and RL \cite{Levin+al:1997,Singh+al:1999}, to methods based on partially observable MDP (POMDP) models (e.g., \cite{williams2007partially}), to more recent approaches adopting deep learning representations (e.g., \cite{li2016deep}). Despite this, deployments of RL ``in the wild''  in large-scale dialogue systems, such as smart assistants like Alexa, Siri or Google Assistant, are rare
(though exceptions exist, e.g., \cite{Serban2017Deep}; see Related Work below). Indeed, building such systems remains a formidable challenge.

Aside from the infrastructure hurdles associated with scalable, real-time systems, there are inherent modeling challenges when building open-ended conversational bots. First, the state space of such bots is massive, even within specific verticals, and care is needed to craft effective state representations for RL algorithms. Second, the action space is also in principle ``unbounded,'' and imposing reasonable limitations on actions comes with its own difficulties, including the fact that the set of candidate actions may vary as the conversation progresses.
Finally, the design of suitable reward functions for open-ended dialogue can be quite subtle. In this work, we rely on crowd-sourced labels.

We present a real-time, open-ended dialogue system that uses RL to power a bot's conversational skill at scale. We address the challenges above using a novel RL construction. We first exploit powerful supervised models---specifically, RNNs and transformers---to provide a succinct embedding of the conversation state. Second, we use the fact that a relatively small set of ``reasonable'' \emph{candidate actions} can be generated at each conversation turn \cite{szpektor2020dynamic}. From an RL perspective, this can be viewed as a stochastic realization of the full action space, so we use an RL approach tailored to such \emph{stochastic action sets} \cite{boutilier2018planning}.
%We also draw a connection to \emph{mixture-of-experts} models \cite{XXX}.
We also explore the use of alternative SOTA RL techniques and training methods, including continuous-action optimization \cite{ryu2019caql}, conservative Q-learning \cite{CQL} and novel off-policy evaluation algorithms \cite{nachum2019dualdice}. 
%We investigate and compare various combinations of RL methods and state representations.

\begin{figure}
\begin{subfigure}{1\columnwidth}
  \begin{center}
  \includegraphics[width=0.9\columnwidth]{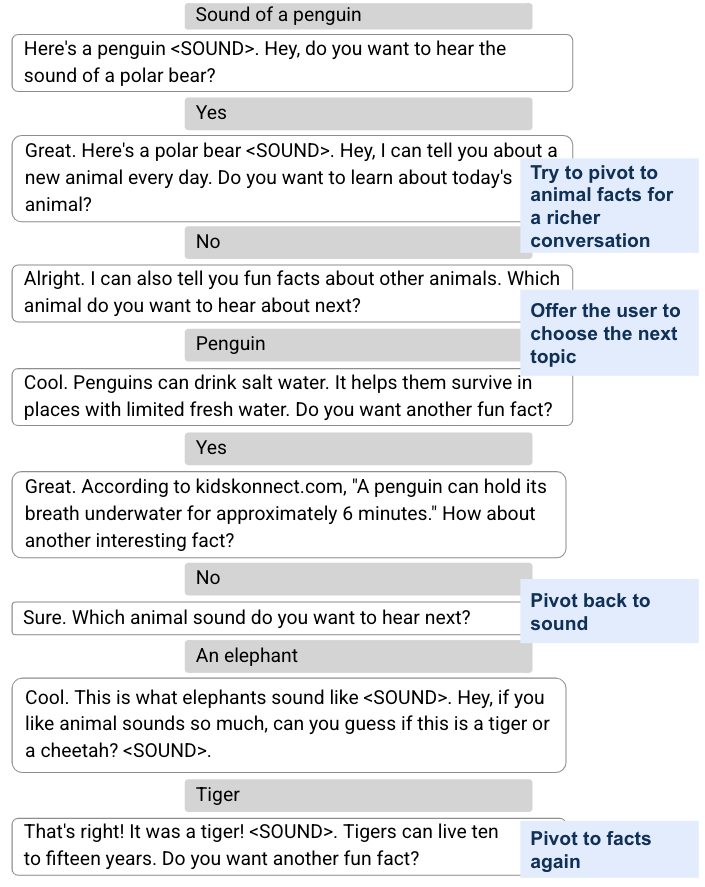}
  \caption{RL model}
  \label{fig:exp_dialogue}
  \end{center}
\vspace{2mm}
\end{subfigure}
\begin{subfigure}{1\columnwidth}
  \begin{center}
  \includegraphics[width=0.9\columnwidth]{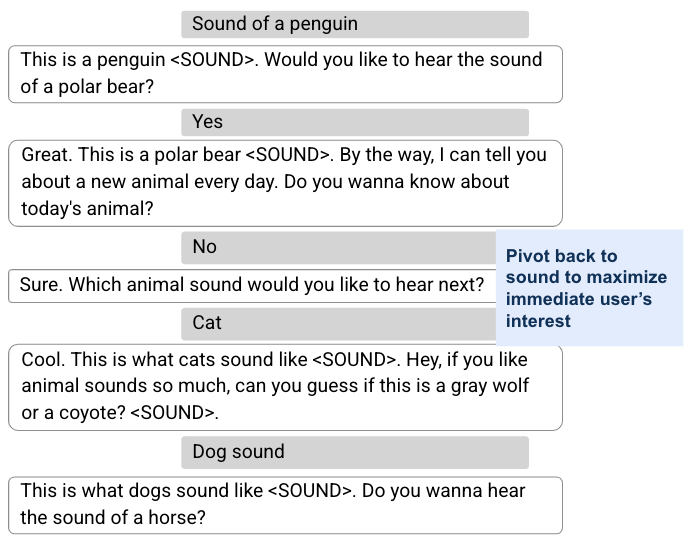}
  \caption{Supervised model}
  \label{fig:ctl_dialogue}
  \end{center}
\end{subfigure}
\caption{Example conversations conducted by (a) an RL model and (b) a supervised model, previously deployed in the Google Assistant, showcasing the rich pivoting of the RL model vs. the supervised approach.}
\label{fig:exp_and_ctl_dialogue}
%\vspace{-4mm}
\end{figure}

We first evaluate our methods using offline data. We then describe the deployment of our system
``in the wild'' in the Google Assistant, specifically as part of the animal domain experience described by \citet{szpektor2020dynamic}. We demonstrate the effectiveness of our RL-based approach at dynamic planning and driving open-ended dialogue: relative to a SOTA non-RL (transformer) baseline, our bot substantially improves a number of key metrics, including conversation length, cooperative responses and explicit positive feedback. We perform a novel and extensive comparison of many RL architectures in real-world settings and generate unique insights into their relative performance. An example dialogue is shown in Fig.~\ref{fig:exp_and_ctl_dialogue}, showcasing the rich pivoting of our best-performing RL model vs.\ the supervised approach. 
Our model, now deployed in the Google Assistant, marks an important milestone in the use of RL for driving real-time, engaging, open-ended, conversations. %Example conversations conducted by our best RL model and by the supervised baseline are shown in Figs.~\ref{fig:exp_and_ctl_dialogue} and \ref{fig:ctl_dialogue}, respectively. While the supervised model aims at maximizing the immediate user's interest, the RL model tries to dynamically plan towards a more diverse and richer experience.

% \begin{figure}
%   \includegraphics[width=1\columnwidth]{figs/exp_dialogue}
% \caption{Example conversation conducted by an RL model.}
%   \label{fig:exp_dialogue}
% \end{figure}

The key ingredients and insights of our approach are threefold. First, we use an effective representation of the RL state, leveraging pre-trained supervised models that encode the conversation history. Second, we limit the action space to a small set of generated candidate actions while allowing multiple actions in a single turn to compose rich bot responses. This granularity decouples content generation and dialogue planning. Finally, we adapt recent state-of-the-art RL algorithms that are well-suited to our dynamic action space to the candidate-generation decomposition we adopt here. 

%Our contributions are as follows: (a) we propose an RL-based dialogue manager (DM) in conversational domain exploration; (b) we introduce and compare several RL architectures, including stochastic action and continuous action modelings as well as RNN and transformer based state representation; (c) we report results and insights of a live experiment of our different models deployed in a commercial assistant.\amirg{will be nice if you can point to "what made it work" and count that as a key contribution}

\section{Related Work}
\label{sec:related}

The use of RL for dialogue dates back more than two decades. Statistical research focuses on task-oriented dialogues and uses MDPs \cite{Levin+al:1997, Singh+al:1999, singh2002optimizing, walker2000application} or POMDPs \cite{williams2007partially, young2010hidden}. \citet{henderson2008hybrid} introduce function approximation to reduce the resulting large slot-filling state space. \citet{casanueva2018feudal} propose a feudal RL model which decomposes the decision by first selecting a subset of primitive actions, then choosing the actual action. These methods each model the state and action spaces using handcrafted semantic representations, such as slots and dialogue acts. This restricts such approaches to simple domains with limited slot-filling.
% state space and limited action space.

More recent work leverages deep neural networks (DNNs) to obviate the need for these so-called summary states and actions \cite{gavsic2011line}. \citet{fatemi2016policy} consider a low-dimensional continuous-valued state space. \citet{liu2018dialogue} encode the dialogue state using an RNN but translate the representation to slot-value pairs. In both cases, the action space is restricted to a small number of dialogue acts and the approaches remain limited to specific task-based domains, with no clear extension to open-ended dialogues.

Building on advances in neural generative models, another line of work applies RL to directly learn a response generation model conditioned on the dialogue history. Actions are often defined at the word level, so the action space is the entire vocabulary \cite{jaques2020human, li2016deep, li2017adversarial, shin2020generating}. This approach suffers from several drawbacks: the action space is very large; word-level RL performs credit assignment poorly at an unnatural level for dialogue planning; and this may affect decoder performance, leading to incomprehensible utterances \cite{zhao-etal-2019-rethinking}.

Related approaches model actions as latent variables, inducing a latent action space, thus decoupling the discourse-level decision making from NLG \cite{zhao-etal-2019-rethinking,saleh2020hierarchical}. However, they focus on specific domains (e.g., price negotiation, slot-filling or chitchat) rather than grounded open-ended dialogues. \citet{Serban2017Deep} proposed MILABOT, as part of the Amazon Alexa Prize competition, where a DM selects a response from several generated candidates. Our approach is similar in spirit, but allows one to combine several candidates in the same bot turn to compose a richer response. This seemingly small difference is vitally important as it allows our RL model to make decisions at an effective granularity. We note that while MILABOT was restricted to an A/B testing evaluation within a competition, our RL model is deployed in the Google Assistant.

\section{Dynamic Composition}
\label{sec:dynamic_comp}

%\gal{I think it is OK to spend a section describing this (let's try and be succinct) but it is important to say a few words at the get go about why these details are important for the understanding of what comes next. Something like the structure of the system defines our action space and is thus importance to the understanding of the resolution in which RL works}

In this work, we build on the \emph{dynamic composition} approach introduced by \citet{szpektor2020dynamic}. This dialogue management model limits the action space using specific \emph{content providers} to propose candidate utterances, which are dynamically selected by the \emph{dialogue manager (DM)}. We adopt this scheme to manage action complexity in our RL approaches.
% , described in Section~\ref{sec:dm_combined}, and is therefore important to their understanding.

\begin{figure*}
  \includegraphics[width=1.8\columnwidth]{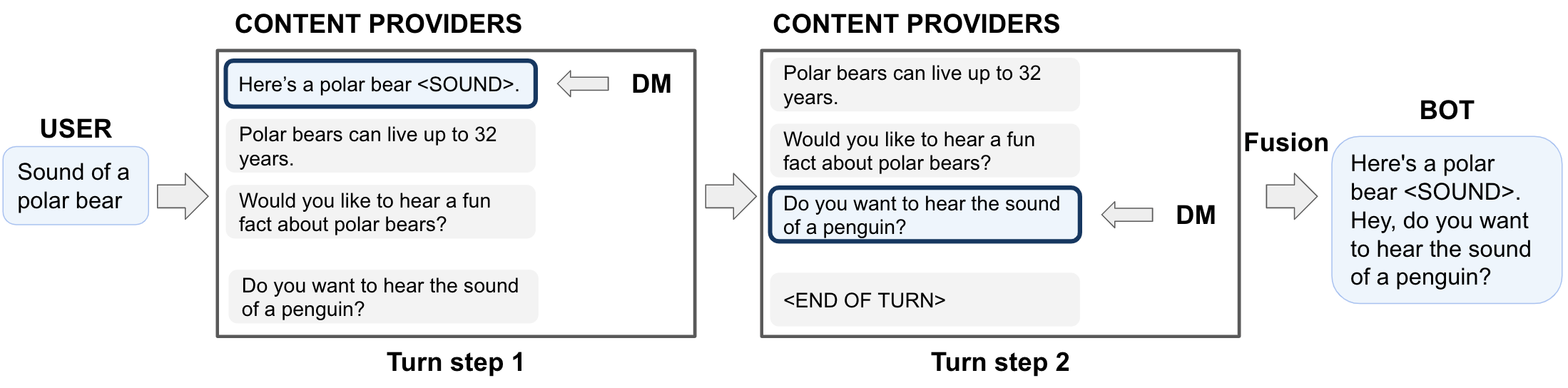}
\caption{Dynamic composition flow diagram. A few candidates from different providers are shown in each step and the one selected by the DM is highlighted in blue.}
  \label{fig:dynamic_comp}
\end{figure*}

Dynamic composition decouples content (utterance) generation from selection. Given candidate utterances proposed by each of several providers, the DM scores and selects a suitable utterance as (part of) a response given the current conversation history. The bot response can be composed of several utterances, which are generated and selected sequentially and dynamically (see Fig.~\ref{fig:dynamic_comp}).
Additional components include an NLU module and a sentence fusion module that merges the selected utterances into a coherent response. We describe each of these components in turn.
%\gal{Going back to previous comment, keep length proportional to the importance for {\bf this} paper. Fusion, for example, can be shortened dramatically without any loss of needed details. One could also argue the same of the NLU}

% \paragraph{\textbf{Natural Language Understanding}}
\vspace{1mm}
\noindent
{\textbf{Natural Language Understanding.}}
% {
At each turn of the conversation, user input is analyzed by an NLU module, comprising two components: a \emph{focus tracker} and a \emph{user answer interpreter}. The \emph{focus} of the conversation is the set of entities \emph{currently} being discussed. %Here, entities may refer to proper nouns such as people and organizations, common nouns such as animals, events such as sport games, and even properties such as height or speed. 
If the last turn ended with a bot question, the \emph{answer interpreter} classifies the user answer w.r.t.\ the question type. For example, the response \emph{types} expected for yes/no questions (e.g., ``Do you want to hear more about cheetahs?''), are `yes' (e.g., ``sure,'' ``I want to hear about cheetahs''), 'no', and ignoring the question; list selection questions %(e.g., "Do you want to catch up on yesterday's animal of the day or hear today's creative challenge?"), semi-open questions
(e.g., ``Which animal do you want to hear about next?'') and quizzes (e.g.,``Can you guess which animal this is, a lion or a tiger?'') include animals as potential responses. Table~\ref{tab:nlu_provider_example} illustrates the concepts of focus and answer interpretation.
% ceb: mentioned below, don't repeat here
% and content provider dialogue acts (the latter described next).
%\todo{Debby}{We should probably give a more concrete description of how these components work.}
% }

% \paragraph{\textbf{Content Providers}}{
\vspace{1mm}
\noindent
{\textbf{Content Providers.}}
% {
Given the conversation history and the selected utterances so far in the current turn, content providers propose candidate utterances w.r.t.\ this context. They rely on different sources (e.g., news, knowledge graph) to extract relevant content, which includes both structured and unstructured data. Structured providers generate text for their candidate utterances using templates, while unstructured content is quoted verbatim with attribution.
Content is expressed in different forms via \emph{dialogue acts} such as statements, preference queries, navigation questions and quizzes. Some of these, referred to as \emph{conversational drivers}, aim to proactively increase user engagement (e.g., focus changing, questions). Examples of such dialogue acts are provided in Table~\ref{tab:nlu_provider_example}.
%for a sample conversation.

\begin{table*}
\begin{small}
\begin{tabular}{lllll}
\multicolumn{2}{c}{\textbf{Dialogue}} &
\multicolumn{1}{c}{\textbf{Focus}} &
\multicolumn{1}{c}{\textbf{User Answer Interp.}} & \multicolumn{1}{c}{\textbf{Dialogue Act}} \\
\hline
\User{}              &  Sound of a polar bear & polar bear & & \\
\hline
\xTalkBot{}              & Here's a polar bear <SOUND>. 
 & polar bear & & Sound \\ 
                        % & Hey, do you want to hear the sound of a penguin?  & penguin & & Focus change \\
                        %  &  & & &  + Yes/no question \\
                        & Hey, do you want to hear the sound of a penguin?  & penguin & & Focus change + Yes/no question \\                         
\hline
\User{}              & Yes & penguin & cooperation &\\
        \hline
\xTalkBot{}              & Great. & penguin & & Ack \\
        & This is a penguin <SOUND>.  & penguin & & Sound \\ 
       & Which animal do you want to learn about next? & & & Open question\\ 
        \hline
\User{}              & Tell me about polar bears & polar bear & cooperation & \\
\hline
\xTalkBot{}              & Cool. & polar bear & & Ack\\  & Churchillwild.com says that "Polar bears will wag & & & \\  & their heads from side-to-side when they want to play". & polar bear & & Fact \\ 
        \hline
        \end{tabular}
\vspace{2mm}
\end{small}
\caption{Illustration of the concepts of focus, user answer interpretation and dialog acts for a sample conversation.}
\label{tab:nlu_provider_example}
\vspace{-2mm}
\end{table*}
% }

% \paragraph{\textbf{Dialogue Manager}}{
\vspace{1mm}
\noindent
{\textbf{Dialogue Manager.}}
% {
In each step of the bot composition loop, providers generate candidates for the next utterance to be appended to the response constructed so far. Given a set of candidates and the conversation context, utterance selection is performed by a learned \emph{DM}. This step is repeated until the DM assesses that the response is a relevant and engaging bot response.
In \cite{szpektor2020dynamic}, the DM is implemented as an RNN encoder, trained in a supervised fashion (see Sec.~\ref{ssec:supervised_dm}). In this work, we develop DMs trained using RL.
% }

% \paragraph{\textbf{Sentence Fusion}}{
\vspace{1mm}
\noindent
{\textbf{Sentence Fusion.}}
% {
The output of the composition loop is a sequence of utterances, which still needs to be fused into a coherent bot response. Simple concatenation of the utterances typically results in cumbersome, unnatural, verbose responses, such as ``On average, male lions weigh 420 lbs. On average, male lions are 3.9 feet tall. That means that lions are about as tall as a piano.'' 
\emph{Sentence fusion} combines the selected utterances into a single cohesive response \cite{barzilay2005sentence,MarsiK05,GevaMSB19}, such as ``On average, male lions weigh 420 lbs and are 3.9 feet tall. That means that they're about as tall as a piano.''
This module uses the following 
% fusion phenomena: 
techniques: 
(a) pronominalization, (b) removal of repetitive context mentions and (c) introduction of a discourse marker between sentences. Our fusion model is based on LaserTagger \cite{malmi2019encode}, a sequence labeling architecture in which each token in the input text is classified to be either copied as-is, deleted or substituted with a phrase taken from a small, predefined vocabulary, typically containing pronouns and connectives.
%which is particularly suitable for text editing tasks in which the output text highly overlaps with the input text. Essentially, LaserTagger is a sequence labeling architecture in which each token in the input text is classified to be either (a) copied as-is to the output sequence; (b) deleted; or (c) substituted with a phrase taken from a small, predefined phrase vocabulary. This phrase vocabulary will usually contain pronouns and connectives such as ``and''.
% }

% \input{dialogue_manager}

% \input{rl_dm}

\section{Reinforcement Learning for the DM}
\label{sec:dm_combined}
Dynamic composition is realized by \citet{szpektor2020dynamic} with supervised training of an RNN-based DM. This limits the construction of the next bot response to be \emph{myopic}, as it is optimized for maximal \emph{immediate} reward.
However, since the main goal of the DM is to conduct complex, engaging multi-turn conversations, it should target the natural complexity of human-to-human conversations, which are typically not conducted in a myopic, turn-by-turn manner, but rather reflect some degree of look ahead and dynamic planning. 
%\gal{We always come back to this point: we need to be careful of putting the focus on a broader agenda since a lesson with predefined steps is also a planned broader agenda but has nothing to do with RL but rather pre-planning. We have to also talk about the need for dynamic re-planning.}
For example, a conversation may comprise several steps leading to an intended goal, such as knowledge transfer; or to make a conversation more engaging, one might intersperse interesting facts throughout, or build tension towards an eventual resolution. Such capabilities require a bot be able to choose responses that lead toward such non-myopic ends, and adapt to user responses/queries by dynamically re-planning its trajectory accordingly.

%Another limitation of the supervised learning approach is the need for a dataset with feedback for each bot response. Such feedback is not available in standard conversational datasets and thus requires a dedicated data collection process using human raters, which is complicated, expensive and may create distributional shifts from real conversations. RL only requires feedback on some of the conversation turns or at the conversation level. This will allow us to extend the training data to further datasets in the future.
%\gal{I wouldn't put something that we don't do now as a core motivation. This is more of a discussion of what the approach will give us in the future. It also dilutes the main story about planning}

To this end, we develop an RL framework for open-ended dialogue that builds on the dynamic composition architecture, and propose a number of concrete instantiations of it. In Sec.~\ref{ssec:mdp}, we formulate the underlying MDP
% \citep{puterman2014markov}
that captures the spirit of the
content-provider decomposition.
% drawing analogies to mixture-of-expert models.
%In Sec.~\ref{ssec:supervised_dm}, we explicate the use of the original RNN encoder \cite{szpektor2020dynamic}, as well as a transformed-based encoder, as our state representation.
% ceb: way too much detail for a section intro. Just repeats what will be said later!
% which characterizes our dialogue management setting and then describe the state representation that we adopt. The state is the concatenation of (i) the output of the dialogue history encoder, either RNN or transformer, (ii) the embedding of the last user input and (iii) context metadata features, including the conversation turn index and the composition loop utterance index.
%We propose a two-step Q-learning approach in Sec.~\ref{ssec:two_steps_rl}, and develop two core Q-learning algorithms---one based on a stochastic action set formulation \cite{boutilier2018planning}, the second using continuous action space \cite{ryu2019caql}---motivated by specific properties of the MDP and our representations. We also develop an end-to-end Q-learning method in Sec.~\ref{ssec:E2E} that does not require the language encoders for state generation. Finally, we consider conservative Q-learning (CQL) regularization \cite{CQL}, which can be applied to each of %these models.
In Sec.~\ref{ssec:supervised_dm}, we discuss the use of the underlying supervised model for state representation. We then propose a two-step Q-learning approach in Sec.~\ref{ssec:two_steps_rl}, with algorithmic variants motivated by specific properties of the MDP and our representations. We also develop an end-to-end Q-learning method in Sec.~\ref{ssec:E2E} that does not require the language encoders for state generation. %Finally, we consider conservative Q-learning (CQL) regularization \cite{CQL}, which can be applied to each of these models.

% ceb: Moved the 2-step figure. It does not belong in the intro. Moved it to Sec.4.3.

%\gal{The following feels a bit mechanical. I think a conceptual outline of the section is needed that described the interplay between the different components}
%To solve this MDP, we propose two RL models. The first, presented in Section~\ref{ssec:two_steps_rl}, is a two step Q-learning approach which relies on state encoding using the supervised architectures from Section~\ref{ssec:supervised_dm}. The second is an end-to-end Q-learning approach in which the state encoder and DM are jointly trained to minimize a Q-learning loss, as shown in Section~\ref{ssec:E2E}. Last, we describe the training data used to train both the supervised and RL architectures (Section~\ref{ssec:data}).

\subsection{MDP with a Stochastic Action Space}
\label{ssec:mdp}

% \gal{Very hard to follow section. Set the stage for the core ideas and consider paragraph headers to make it easier to follow. Some of that appears in the paragraphs that follow but only after opening with a bombardment of notation}
% \todo{yinlam}{Simplified and added a header}

We begin with an MDP formulation of the DM problem upon which our RL methods operate.
%To explicitly model the ability to plan so as to maximize expected cumulative rewards (given for desired behavior), we propose an RL approach, which allows planning based on multiple sequential decisions and cumulative expected return (e.g., pivoting to other entities, asking questions).
% Our MDP is $M = (\mathcal X, \mathcal A, P, R, \beta, \gamma)$, 
We assume a state space $\mathcal X$, action space $\mathcal A$, transition kernel $P$,  reward function $R$, initial state distribution $\beta$ and discount factor $\gamma$,
and aim to optimize the cumulative discounted return $J(\pi):=\mathbb E[\sum_{t=0}^{\infty}\gamma^t r_t\mid P,R,\beta,\pi]$ which captures the long-term value of the conversation. The RL DM \emph{policy} $\pi$ is an action distribution conditioned on state $x\in\calX$. An optimal DM $\pi^*$ is found by solving $\max_{\pi \in \Pi} \,J(\pi)$, where $\Pi$ is the space of all DM policies. We discuss each of these elements in turn.

% \todo{Debby}{IMO, we should make it clear at this stage that the following is not the case in our approach. }
% \todo{yinlam}{Reworded. WDYT?}

Much RL research in dialogue management defines the state and action spaces to be the tokenized language space
\cite{li2016deep,jaques2019way,asadi2016sample}. For instance, the state $x$ is the tokenized user-agent conversation history to that point, while an action $a$ is the DM's output sentence (generated token by token).
However, since the state and action spaces in this formulation are both combinatorial, even with a medium-sized vocabulary the corresponding tokenized spaces  grow exponentially, thus making RL intractable.
% % (that is different from our proposed approach)
% is to have the state space $\mathcal X$ as the tokenized language space, state $x$ as the current user-agent tokenized conversation history, and initial state $\beta$ as the initial user's (language) input query. The action space $\mathcal A$ is also the tokenized language space with action $a$ being the DM's output sentence (generated token by token). Transition kernel $P$ models the user's response to the DM's output, and reward $R$ is a rater's score that measures the quality of the current bot response. 
% However, since the state and action spaces in this general DM MDP are both combinatorial, even with a medium-sized vocabulary token dictionary the corresponding tokenized language spaces will grow exponentially, thus making RL intractable in this setting.
We handle the combinatorics of state space by leveraging state-of-the-art language models, such as RNNs or transformers, to encode conversation history $x$ with a $d$-dimensional embedding $\phi_x\in\mathbb R^d$ (see Sec.~\ref{ssec:supervised_dm} for details), thus replacing the large discrete state space by the continuous embedding space $\mathbb R^d$.

We also differ from typical RL dialogue models in our treatment of the action space. Rather than a generative language model that directly outputs sentences, we leverage the dynamic composition framework (Sec.~\ref{sec:dynamic_comp}) to render the action space tractable. Specifically, at any turn, each content provider proposes a small set of utterances. This \emph{dynamic-composition, content-provider (DCCP) action decomposition} ensures the DM policy need only score and select from a (relatively small) 
% ceb: we can quantify it if we're prepared to make assumptions: (MC choose k) if the response has k utterances, we have M providers, each of which sends C candidates. Depends on whether we can choose mulitple from the same provider or not, etc.
discrete set $\calA_x$ at state $x$, i.e. the set of candidate utterances.
%. We further scale down the action space by treating each \emph{utterance} inserted into the composed bot response as our target action (as well as the decision to end a response and move to fusion).
Note that by working at the utterance level rather than at the level of the full bot response (a fused concatenation of K such utterances), we remove the small-scale combinatorics of the action space, at the cost of extending the horizon of the RL problem. But this does not sacrifice the optimality of the policy.

Importantly, since the providers use potentially arbitrary logic to generate candidates, the realization of $\calA_x$ may differ with each occurrence of $x$. This puts us in the realm of non-standard MDPs with \emph{stochastic action sets} \cite{boutilier2018planning}. Fortunately, we consider Q-learning methods below that handle this directly.\footnote{We note that the DCCP action decomposition is related to certain hierarchical approaches to RL, such as feudal RL \cite{dayan-hinton:1996,casanueva2018feudal}, and to mixtures-of-experts \cite{jacobs-jordan-nowlan-hinton:1991,kotsiantis2003mixture}, though our DM RL methods do not influence the training of the providers.}

The training data, further described in Sec.~\ref{ssec:data}, is composed of crowd-sourced conversations, generated with a supervised DM. The human evaluators provide a rating for each selected utterance, which are used as rewards that measure the rater's immediate value of the action given the conversation history. As we shall see, the RL model is able to leverage these to learn dynamic planning strategies to improve user engagement.

% ceb: Removing MoE discussion: please see my email. But replaced it with a brief footnote.
% and consider the \emph{mixture-of-expert (MoE)} setting \cite{casanueva2018feudal, kotsiantis2003mixture}, where at each step there exists several content providers that generate a (stochastic) number of candidate sentences $\mathcal A(x)$ based on current state $x$, and our DM is restricted to selecting one as the output action.
% \todo{Debby}{I'm not sure we should introduce an additional notation for the experts/content providers. I would stay at the candidate actions level since the DM works at that level.}\todo{yinlam}{removed $\mu$ notations}

% Therefore, instead of solving the RL problem over a combinatorial action space $\mathcal A$, we transform this problem into \emph{RL with stochastic action space}, whose goal is to find the optimal mixture distribution over candidate actions (MoE policy) $\lambda$, which is an action distribution over the candidate action set $\mathcal A(x)$ at each state $x$. 
% Specifically, let $J^\prime(\lambda)=\mathbb E[\sum_{t=0}^{\infty}\gamma^t r_t\mid P,R,\beta,\lambda]$ be the MoE discounted return, where $P$ and $R$ are the transition kernel and immediate reward functions induced by candidate actions in $\mathcal A(\cdot)$.
% Finding the MoE DM model then becomes (i) solving for the MoE policy $\lambda^*\in\arg\max_{\lambda\in\Delta} \,J^\prime(\lambda)$ and (ii) constructing the MoE DM $\pi_{\lambda^*}(a|x)= \lambda^*(a|x)$, for $a\in\mathcal A(x)$ and $\pi_{\lambda^*}(a|x)=0$ otherwise. 

In the existing dialogue RL literature \cite{li2016deep,li2009reinforcement,chandramohan2010sparse} most algorithms are based on policy-gradient methods \cite{sutton1999policy} because (i) learning an optimal state-action value function with a combinatorial action space is intractable, and (ii) the resulting DM is  a sequence-to-sequence policy model that generates bot responses. The simplification of the action space afforded by our DCCP reformulation allows us to use \emph{value-based} Q-learning \cite{mnih2013playing,boutilier2018planning}, a relatively uncommon approach in large-scale dialogue systems.

% CEB: I DON'T THINK WE NEED THIS
% In a discounting MoE MDP, $Q$-learning aims to learn the fixed-point solution
% of the Bellman operator $F[Q](x,a):=R(x,a) + \gamma\sum_{x'\in\mathcal X} P(x'|x,a)\max_{a'\in\mathcal A(x')}Q(x',a')$ over state-action value function $Q$ because \citep{puterman2014markov} shows that this solution is unique and equal to the optimal Q-function $Q^*(x,a)=\mathbb E[\sum_{t=0}^\infty\gamma^t r_t| P, R, x_0=x,a_0=a,\lambda^*]$. After learning the optimal $Q$-function of this MDP, one can then construct an optimal MoE policy
% $
% \lambda^*(a|x)=\mathbf 1\{a=a^\star(x)\}, \,\text{where}\, a^\star(x)\in\arg\max_{a\in\mathcal A(x)} Q^*(x,a).
% $ 

\subsection{Supervised Model as a State Encoder}
\label{ssec:supervised_dm}

To encode the conversation history into a $d$-dimensional embedding vector, we consider two supervised learning architectures: an improved version of the RNN model from \cite{szpektor2020dynamic} and a transformer-based approach using a pre-trained BERT model. 

% \paragraph{\textbf{Supervised RNN Architecture}}{ 
\vspace{1mm}
\noindent
\textbf{Supervised RNN Architecture.}
We modify the two-level hierarchical RNN encoder \cite{szpektor2020dynamic} by replacing the first-level gated recurrent unit (GRU), that encodes the user and bot utterances, by a pre-trained sentence embedding module \cite{USE}, which is fixed during training. 
% Our experiments show that 
These provide a clear advantage over the first-level GRU, which we attribute to training on a large, general corpus.
The resulting sentence embeddings are fed to a GRU along with non-textual metadata features, including: the conversation turn index, the candidate dialogue act, the number of tokens in the constructed response, whether the candidate offers to change the focus.
% }

% \paragraph{\textbf{Supervised Transformer-based Architecture}}{
\vspace{1mm}
\noindent
\textbf{Supervised Transformer-based Architecture.}
Our second supervised model uses a transformer architecture \cite{Transformer}. We consider two variants: a text-only BERT model and a combined text-metadata model. In both, the input is the concatenated sequence of user and bot utterances (i.e., conversation history) and a candidate utterance to be scored.
% delimited by the \texttt{[SEP]} token.
In the second, we concatenate the per-token contextual representation vectors produced by the BERT model and an embedded representation of the metadata features for the utterance of which the token is a part. The resulting concatenated vectors are fed into another small transformer. Experiments on manually annotated data show the text-only variant outperforms the second, corroborating the hypothesis that the transformer's better use of the input text---specifically, the ability to attend to the history when processing candidates---obviates the need for 
% construction and engineering of 
additional features.
% }

% \paragraph{\textbf{State Representation}}{
\vspace{1mm}
\noindent
\textbf{State Representation.}
Our state representation uses the output of the dialogue-history encoder, either the RNN hidden state or the pooled output of the last transformer layer.
% (corresponding to the \texttt{[CLS]} token).
The state is constructed by concatenating this encoding with the sentence embedding of the last user input and  context features, e.g., the conversation turn index and the composition turn step index.
% }

%\todo{Debby}{The RL state is actually the concatenation of the RNN/transformer state with the last user input, the current bot response and corresponding metadata. Should we get into these details?}
%\gal{I think the RL state is a core issue in this paper so this is one of the topics on which I would take it slowly and not try to save space}

% \todo{ceb}{Please move this paragraph, it does not belong under "State Repn." Put at the top of the whole section, or embed it in the next subsection.}
% In the following we consider two RL DM approaches, which rely on Q-learning. The first is a two-step model in which the state is first encoded either by the pre-trained RNN or transformer before solving for the MoE DM model. The second is an end-to-end (transformer) approach in which the encoder and MoE DM are jointly trained to minimize the RL loss.
% The loss is minimized over mini-batches $B$ of data $(x, a, r, x')$ sampled from the offline conversation history replay buffer $\mathcal B$ \citep{zhang2017deeper}.

\subsection{The Two Step Q-model Architecture}
\label{ssec:two_steps_rl}

% \gal{Need to cut down drastically, impossible to understand the mid-high level setting. It is OK to spend a little more time on stochastic actions while explaining why we do so but even on that one we need to focus more on the essence of the method rather than all the details. What you really want the reader is to understand {\bf Why} we chose to try each method plus basic details}
% \todo{yinlam}{updated, WDYT? I tried to just have the technical details enough to describe the main training loss and put more descriptions on motivations}

\begin{figure}
  \includegraphics[width=1\columnwidth]{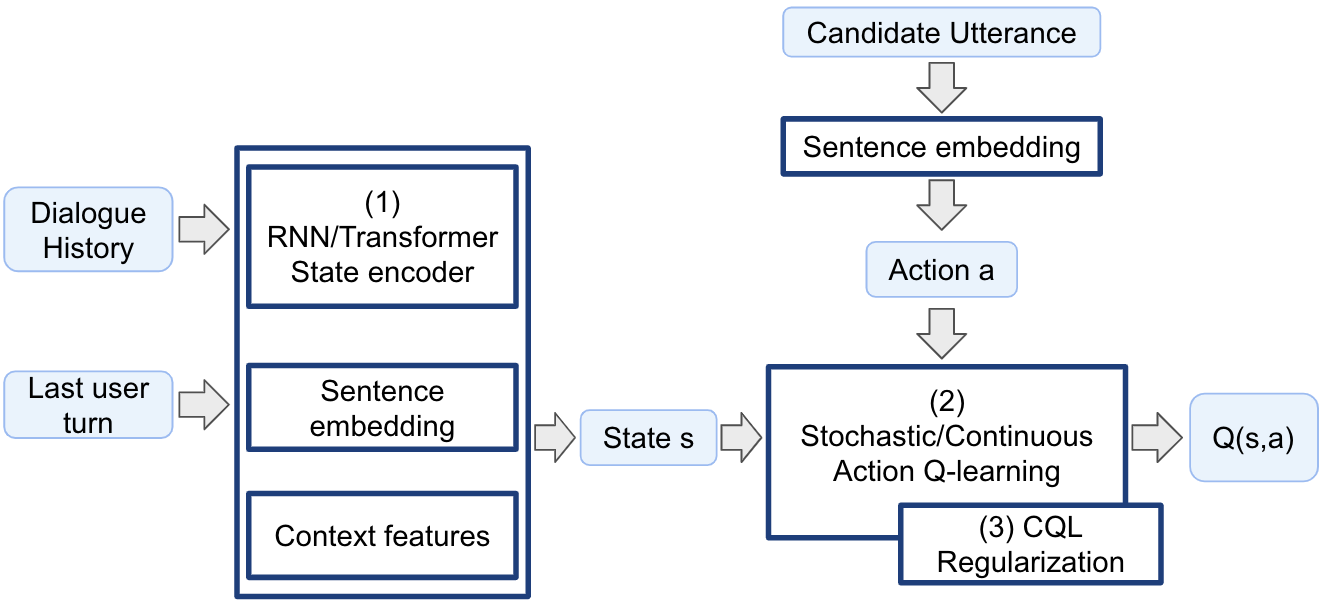}
\caption{Two-step Q-learning schema. The state is the concatenation of (i) the output of the dialogue history encoder, either RNN or transformer (denoted by (1)), (ii) the embedding of the last user input and (iii) context features, including the conversation turn index and the composition turn step index. The action is represented by its embedding. We consider both stochastic action and continuous action Q-learning approaches (denoted by (2)) potentially with the added CQL regularization (denoted by (3)).}
  \label{fig:rl_models_schema}
%\vspace{-2mm}
\end{figure}

We now develop several RL approaches for the DM which rely on Q-learning. Our first
approaches use a \emph{two-step model} in which the state is first encoded by a language model (either a pre-trained RNN or transformer) before being passed to the DM policy.
Figure~\ref{fig:rl_models_schema} illustrates how these building blocks come together in the two-step approach.
Given a pre-trained state encoder $\phi:\mathcal X\rightarrow\mathbb R^d$ and a sentence encoder $\psi:\mathcal A\rightarrow\mathbb R^h$, we apply two different Q-learning techniques using the encoded state space (using $\phi_x$ rather than $x$) and action space (using $\psi_a$ rather than\ $a$).

% \paragraph{\textbf{Stochastic Action Q-learning (SAQL) \cite{boutilier2018planning}}}{
\vspace{1mm}
\noindent
\textbf{Stochastic Action Q-learning (SAQL) \cite{boutilier2018planning}.}
Our first RL technique applies Q-learning directly to the discrete, stochastic action sets $\calA_x$ as determined by the DCCP decompostion. We adopt the general deep Q network (DQN) approach \citep{mnih2013playing}, using a DNN to represent the Q-function.
Specifically, $Q_{\theta}:\mathbb R^d\times\mathbb R^h\rightarrow \mathbb R$ is a feed-forward DNN with parameters $\theta$, which represents the cumulative discounted value of taking action (or bot utterance) $\psi_a\in \mathbb R^h$ in state (i.e., conversation history encoding) $\phi_x\in \mathbb R^d$.

We train the model using batched conversation data of the form
$B=\{(\phi_{x_i},\psi_{a_i},r_i,\phi_{x_i'}, \calA_{x'_i})\}_{i}$ (see Sec.~\ref{ssec:data} on data generation), where the $i$th data point includes:
the embedded state $\phi_{x_i}$ and action $\psi_{a_i}$ at some conversation turn, the observed next state $\phi_{x_i'}$, the realized reward $r_i$, and the set of actions $\calA_{x'_i}$ made available by the content providers at the next state. This latter component is non-standard in DQN, but necessary due to the stochastic nature of the action sets---training the Q-function (see below) exploits maximization over the \emph{realized} set of candidate actions generated by our providers at the next state, an approach shown to be sound in \cite{boutilier2018planning}.

We learn $\theta$ by minimizing the \emph{mean squared Bellman error}:
$$\min_\theta \sum_{i=1}^{|B|}(Q_\theta(\phi_{x_i},\psi_{a_i})-r_i-\gamma \max_{a'\in\mathcal \calA_{x'_i}}Q_{\theta^{\text{target}}}(\phi_{x'_i},\psi_{a'}))^2,$$ where $Q_{\theta^{\text{target}}}$ is a \emph{target} $Q$-function, used to improve training stability in DQN \cite{mnih2016asynchronous} (note the use of
the realized action set $\calA_{x'_i}$ in the maximization).
%
% ceb: I'm not sure we need to get into these details. Referring to DQN suffices. If someone knows DQN, they don't need this. If they don't know DQN, this won't help.
%
% The motivation of having a target $Q$-function is to improve training stability. According to \citep{mnih2016asynchronous},  instead of training these weights jointly, $\theta^{\text{target}}$ is updated in a separate iterative fashion using the previous ${\theta}$ for a fixed number of training steps, or by averaging $\theta^{\text{target}}\leftarrow \tau\theta+ (1-\tau) \theta^{\text{target}}$
% for some small momentum weight $\tau\in[0,1]$.
%
Under this loss, RL is $\ell_2$-regression of $Q_\theta$ w.r.t.\ target labels $r+\gamma \max_{a'\in\mathcal \calA_{x'_i}}Q_{\theta^{\text{target}}}(\phi_{x'},\psi_{a'})$, which tries to match the value function $Q_\theta$ with its Bellman backup.

% $F[Q_{\theta^{\text{target}}}]$, so that when these two quantities are close to each other, then according to the fixed-point theorem \cite{puterman2014markov} the corresponding solution $Q_{\theta^*}$ is a good approximation of the optimal value function.

% ceb: seems unnecessary?
% Recall that the presence of a finite stochastic candidate action set $\mathcal A(x')$ at each next state $x'$ is a key component in the MoE formulation that leads to a tractable $Q$-learning algorithm. At almost all data points the size of the candidate action set $|\mathcal A(x')|$ is much smaller than that of the original language action space $|\mathcal A|$, therefore even by exhaustive enumeration solving the inner maximization problem to compute the target label 
% becomes rather tractable. 

% For more stable training, we further augment DQN using \emph{double Q-learning} ~\citep{hasselt2010double}, whose loss is $\min_\theta \sum_{i=1}^{|B|}(r_i+\gamma Q_{\theta^{\text{target}}}(\phi_{x'_i}, \arg\max_{a'\in\mathcal A(x')} Q_{\theta}(\phi_{x'_i}, \psi_{a'}))-Q_\theta(\phi_{x_i},\psi_{a_i}))^2$. 
We refer to this approach as \emph{stochastic action Q-learning (SAQL)} to reflect the stochastic action sets used in training. Once SAQL converges, the DM policy is
$
\pi^\ast(x) \in \arg\max_{a\in\calA_x} Q_{\theta^*}(\phi_x,\psi_a),
$. That is, at inference time, the $Q$-model is applied to each candidate action, and the DM responds with the action with the greatest Q-value given the current dialogue state.
% }
% ceb: why are we even discussing stochastic policies? we don't use them
% $
% \lambda_{\theta^*}(a|x)=\mathbf 1\{a=a^\star(x)\}, \,\text{where}\, a^\star(x)\in\arg\max_{a\in\mathcal A(x)} Q_{\theta^*}(\phi_x,\psi_a),
% $ i.e., at inference time the $Q$-model is applied to each candidate action, and the DM selects the one with the highest value to output.
% % }

% \paragraph{\textbf{Continuous Action Q-learning (CAQL) \cite{ryu2019caql}}}{
\vspace{1mm}
\noindent
\textbf{Continuous Action Q-learning (CAQL) \cite{ryu2019caql}.}
In the SAQL formulation, action maximization takes place over the discrete set of candidates $\calA_x$. However, the embedding representation means that the action space can also be treated as continuous, and we can consider maximization over this wider space. Continuous-action RL problems are common in areas like robotics \citep{kober2011policy}, and typically policy-gradient algorithms are used to  learn a return-maximizing policy \citep{sutton1999policy,silver2014deterministic}. However, such methods are often data-inefficient and impractical when faced with high-dimensional action spaces, both issues present in dialogue systems. Instead, we consider the use of \emph{continuous action Q-learning (CAQL)} \cite{ryu2019caql} to solve the continuous-action variant of our DM policy.

Roughly speaking, using CAQL, when faced with a next state $x'$ while training the Q-function $Q_\theta$, we do not restrict ourselves to maximizing over the discrete action set $\mathcal{A}(x')$, but instead maximize over the entire embedding space $\psi$, minimizing:
% . Specifically, we minimize the following loss: 
$$\min_\theta \sum_{i=1}^{|B|}(r_i+\gamma Q_{\theta^{\text{target}}}(\phi_{x'_i}, \arg\max_{\psi} Q_{\theta}(\phi_{x'_i}, \psi))-Q_\theta(\phi_{x_i},\psi_{a_i}))^2.$$
This approach has advantages over $SAQL$: one need not record the realization $\calA_{x'}$ of the stochastic action sets in the data set, and continuous action maximization (see below) can be more effective when the set of candidate actions (utterances) is moderate or large in size. However, CAQL will generally overestimate the true value of its policy, since it hypothesizes the use of embedded actions that are never generated by any content provider. Indeed, once $Q_\theta$ is trained using CAQL, we \emph{restrict the realized policy} to scoring (and using) only provider-generated candidate actions at inference/serving time.

When $Q_\theta$ is represented by a DNN, the inner maximization is typically differentiable and non-convex. This can be solved optimally for certain classes of DNNs using a mixed-integer program or a first-order method such as gradient ascent (GA) \cite{ryu2019caql}.
% \citep{nocedal1998combining}, which finds the (local) optimum of a differentiable objective function.
We use GA in this work: starting from
an initial
embedded action $\psi^{(0)}$, the optimal embedded action $\arg\max_\psi Q_\theta (\phi_{x'}, \psi)$ is computed iteratively by $\psi^{(t+1)} \leftarrow \psi^{(t)} + \epsilon_{\text{GA}}\nabla_{\psi}Q_{\theta}(\phi_{x'},\psi)|_{\psi=\psi^{(t)}}$, 
where $\epsilon_{\text{GA}}>0$ is a tunable step size.
% This process repeats until convergence or a maximum iteration count is reached.   

% ceb: overestimation is mentioned above Perhaps the stability and bias points below should be worked in as well somewhere.
% While CAQL does not require the candidate action set $\mathcal A(x')$ for training, (in modulo to optimization errors) its max-Q label is an upper bound of that in SAQL. This implies that inherently its Bellman target label is more biased (w.r.t. overestimation) than that of SAQL, which makes CAQL more susceptible to distributional shifts between data and the current policy and more unstable during training.

% }

% \paragraph{\textbf{Conservative Q-learning (CQL)}}{
\vspace{1mm}
\noindent
\textbf{Conservative Q-learning (CQL) \cite{CQL}.}
Our DM problem is an application of \emph{offline RL}, where a model is learned using previously collected
% static data (of
user-bot conversation with no further (online) interaction. Offline RL is prone to overestimation errors induced by the distributional shift between the offline data and that generated by the learned policy \cite{wu2019behavior}. 
This is especially problematic if certain bot actions are rare in the offline data, making their learned $Q$-values very noisy.
To alleviate this, we can apply \emph{conservative $Q$-learning (CQL)} \cite{CQL},
a regularization scheme which learns a ``conservative'' $Q$-function that lower bounds the true Q-function.
% even at points that are not constrained by the Bellman loss. 
CQL can be applied to both SAQL and CAQL (we illustrate it only for SAQL).  

In CQL one augments the Q-learning loss with a behavior regularizer: $\min_\theta \sum_{i=1}^{|B|}
\alpha(\mathbb E_{a\sim \mu}
[Q_\theta(\phi_{x_i}, \psi_a)] - \mathbb E_{a\sim \pi_\beta}
[Q_\theta(\phi_{x_i}, \psi_{a})])+(r_i+\gamma Q_{\theta^{\text{target}}}(\phi_{x'_i}, \arg\max_{a'\in\mathcal A(x'_i)} Q_{\theta}(\phi_{x'_i}, \psi_{a'}))-Q_\theta(\phi_{x_i},\psi_{a_i}))^2$, where 
$\pi_\beta$ is a behavior policy (DM) that approximates the data-generation policy,\footnote{In our setting, the behavior policy $\pi_\beta$ is simply the supervised DM model.} $\alpha > 0$ is a tunable regularization parameter, and $\mu$ is the target policy to be learned. Intuitively, CQL regularization minimizes the differences in Q-values of actions generated by our learned RL DM policy and the behavior (training-data generating) policy. 
% Therefore, besides picking $\mu$ to be a uniform distribution, one better choice is to have
We use target $\mu(a|x)\propto \exp( Q_\theta(\phi_x,\psi_a))$, which corresponds to the optimal policy of entropy-regularized Q-learning \cite{schulman2017equivalence}.

%Theorem 3.3 of \cite{cql} shows that with CQL regularization, the learned Q-function is always a lower bound of the optimal state-action value function under target distribution $\mu$. As $\mu$ approaches the optimal policy, this regularization thus prevents overestimation of Q values in the Bellman backup, which occurs mainly at unseen, ``maximal'' actions.
% }

\subsection{End-to-end Architecture}
\label{ssec:E2E}
% \gal{Contrast this with section 4.3 as a reminder. Motivate why we are also trying that}
% \todo{yinlam: added some motivation sentences}

We now outline an end-to-end (E2E) RL approach that jointly trains the language encoder and the $Q$-function. In contrast to our two-step approaches, by not constraining the DM to using a pre-trained encoder, E2E RL can tune the encoder (hence its representations) to the dialogue task at hand.
This approach is similar in spirit to the original DQN model \cite{mnih2013playing}, in which the $Q$-network consists of both a convolutional DNN that encodes pixel frames (states) and a feed-forward NN that learns the Q-values. 

To learn the $Q$-function in E2E fashion, we apply DQN to $Q(x,a)=Q_\theta(c(x,a))$, where $c(x,a)$ is the concatenation of the conversation history and the current candidate action,
% (separated by a dedicated token denoting the beginning of a bot response); 
and $Q_\theta: \mathcal{X} \rightarrow \mathbb{R}$
is a trainable language encoder (e.g., a transformer \emph{initialized} with pre-trained weights), followed by a feed-forward DNN.
% by a fully-connected layer with $tanh$ activation.
This $Q$-model jointly encodes the raw input conversation and assigns a $Q$-value to each candidate action. This allows us to learn the $Q$-function E2E, without relying on fixed pretrained language encoders. Specifically, with target network $Q_{\theta^{\text{target}}}$ updated as above, in E2E learning we train the $Q_\theta$ by minimizing the mean squared Bellman error. We  use SAQL and formulate the inner maximization as $\max_{a'\in\mathcal A(x')}Q_{\theta^{\text{target}}}(x', a')$.
% at each next state $x'\in\mathcal X$. 
% To alleviate the issue of $Q$-value overestimation bias,
%We can also apply CQL to E2E $Q$-learning if %desired.
% to constrain the $Q$-function to always be the lower bound of its true value.

% This results in the following loss:
% \begin{align*}%\label{eq:dqn_E2E_l2}
% \begin{split}
% \min_\theta \sum_{i=1}^{|B|} &\left[ \left( Q_\theta(x_i,a_i)-r_i-\gamma \max_{a'\in\mathcal{A}(x'_i)}Q_{\theta^{\text{target}}}(x'_i,x') \right)^2 \right. \\
% &+ \left.\frac{\alpha}{|\mathcal{A}(x'_i)|}\sum_{a' \in \mathcal{A}(x'_i)} Q_\theta(x'_i,a') \right]
% \end{split}
% \end{align*}
% where $\alpha\ge0$. 

%\gal{Shouldn't this belong to evaluation as a way to allow us to understand the graphs? I would mention the essence of this (without the numerous names) at the beginning of entire section 4 and then enumerate them when you need to when needed}
%To this end, we denote by (i) SAQL-RNN, CAQL-RNN, SAQL-Transformer, and CAQL-Transformer, the $Q$-models trained with the two-step RL approaches SAQL and CAQL using RNN and Transformer encoders respectively, (ii) SAQL-CQL-RNN, CAQL-CQL-RNN, SAQL-CQL-Transformer, and CAQL-CQL-Transformer the Q-models trained with the above approaches with CQL regularization, and (iii) SAQL-CQL-E2E the Q-model trained with E2E Q-learning with CQL regularization.  

\subsection{Training Data}
\label{ssec:data}
The DM models are trained on crowd-sourced data, generated by human evaluators. Each evaluator converses with the bot until the dialogue derails or comes to a natural end. They then rate the bot responses, assessing each utterance in the composition loop, including those selected and unselected by the DM. Although evaluators were provided with a set of guidelines for assessing bot response quality, the resulting data is noisy and some level of rater-specific subjectivity is included in the ratings.

A dozen evaluators generated $\sim$20K conversations with an average of $3$ bot responses, each with $1$ to $4$ utterances, and with up to $30$ candidates per utterance. For the supervised models, this results in $\sim$1.5M training examples, as each (selected and unselected) candidate corrresponds to a training example. By contrast, RL models only use the labels on selected candidates, giving $150$K labels.

Each candidate utterance is rated on a scale of -3 to 7, with no 0 rating made available. The negative ratings reflect candidates that do not reply to a user question, are out of context, or repeat content that was already mentioned in the conversation. The positive scores correspond to candidates that fit the conversation context well.
%Figure~\ref{fig:xtalk_rater_ui} in the Appendix shows the human-rater UI.

\section{Initial Offline \& Online Evaluation}
\label{sec:offline}
%\gal{Don't forget to mention that the exciting real life experiment is upcoming next}
Before deploying our models in live experiment, we conducted preliminary evaluation of our RL-based DM policies. We describe both (i) \emph{off-policy} counterfactual inference \cite{thomas2015high,li2012unbiased} evaluation and (ii) \emph{on-policy} human (rater) evaluation. Off-policy evaluation can be performed on the existing datasets used to train our models (in our case, generated with the supervised DM). While often easier---hence especially useful for initial model development and tuning---it is less reliable than on-policy evaluation. So we use both methods.
% Therefore, we start with off-policy evaluation to tune our models before generating conversations for each model for on-policy evaluation.

% \paragraph{\textbf{DM Models}}{
\vspace{1mm}
\noindent
\textbf{DM Models.}
We evaluate the following variants of our SAQL and CAQL algorithms, using either the supervised RNN or transformer for state representation, with or without CQL regularization. SAQL-RNN, CAQL-RNN, SAQL-Transformer, and CAQL-Transformer are the Q-learned models trained with the two-step RL approaches SAQL and CAQL using RNN and transformer encoders, respectively. SAQL-Reg-RNN, CAQL-Reg-RNN, SAQL-Reg-Transformer, and CAQL-Reg-Transformer are the same Q-learned models trained with CQL regularization. SAQL-Reg-E2E denotes the E2E Q-learned model with CQL regularization.

The RNN architecture includes a GRU layer with 200 units. The supervised transformer model uses the publicly-available, pre-trained \texttt{BERT-Medium} checkpoint.\footnote{\url{https://github.com/google-research/bert}} The training regime roughly follows that for BERT.\footnote{\url{https://github.com/google-research/bert/blob/master/optimization.py}} For RL models, we use a discount factor $\gamma=0.95$ for SAQL, $\gamma=0.9$ for CAQL and CQL Reg, and $\gamma=0.8$ for E2E RL,\footnote{The choices of these discount factors are mainly to ensure training stability.} and a fully connected feedforward DNN for the $Q$-function. Hyper-parameters 
% for all models
are provided in the appendix.
% }

%\gal{Give the reader a 1-2 sentence reminder of the setup, e.g. supervised model used for state representation of our RL model. Then, just mention the basics of the model (RNN with GRU layer of 200 units, BERT-medium) and move everything else to appendix. E.g., good to say things like training regime roughly follows one for BERT but no need for the other details}

%\gal{It is going to be very hard for the reader to follow the table with names that just differ by a letter at times. I suggest to have table with the 'S' and 'C' models divided by a double line and a column describing the partition so that it will be easier to follow: e.g. 'Non RL', 'Stochastic Actions', 'Continuous Actions'. This will also allow you to use simpler names and a shorter list when mentioning the models that we have. Yes 'Reg' instead of 'CQL' for the regularization.}

%\gal{A couple of sentences of the plan and why you want both on-policy and off-policy would help before starting}

% \paragraph{\textbf{DM Off-policy Evaluation }}{
\vspace{1mm}
\noindent
\textbf{DM Off-policy Evaluation.}
% \todo{debby}{I would drop the abbreviation since it's a bit confusing with on-policy eval.}
% The main goal of OPE is to evaluate the (RL) performance of the DM before its deployment \gal{Isn't that the point of all of the offline evaluation?}. Unlike myopic evaluations, since the sequential conversations generated by interacting with the DM can follow a very different distribution than that of the offline data, one cannot compute the DM's performance simply with Monte Carlo methods.
% \gal{I think this is too fast for those not familiar with the nuances here and on the other hand you have too many math details later. Instead, take it more slowly, avoid difficult sentences (e.g. double i.e.) and cut down on the detailed afterwards}
The main goal of off-policy evaluation is to assess the performance of our RL-based DM policy using existing conversational data.
% specifically, the offline data used to train the policy.
Unlike (supervised) myopic models, RL requires evaluating the reward of full trajectories. Since an RL-based DM can drive sequential conversations that follow a very different distribution than that of the training data,
\emph{off-policy correction} using propensity scoring or related methods is needed \cite{thomas2015high}.
%
% We hereby apply the Dual Stationary Distribution Correction Estimation (DualDICE) \cite{nachum2019dualdice} method to estimate the its performance, which effectively computes the {\em stationary distribution correction ratio}, i.e., the likelihood of the DM-user conversations, i.e., the full dialogue generated by both the DM and the users who respond to that, normalized by the probability with which the conversations appear in the offline data. 
We use \emph{DualDICE} \cite{nachum2019dualdice} for this purpose, a recent SOTA method for off-policy estimation of RL policy values that directly estimates the \emph{stationary distribution correction ratio}, i.e., the ratio of the steady-state probabilities of specific state-action pairs $(\phi_{x_i},\phi_{a_i})$ generated by the RL policy $\pi$ and the data-generating (or behavior) policy $\pi_B$ (which can be estimated from the training data). We provide a high-level overview.

Given a batch $B=\{(\phi_{x_i},\psi_{a_i},r_i,\phi_{x_i'})\}_{i}$ of (embedded conversation history) training data and a DM policy $\pi$, DualDICE learns a feed-forward DNN $\nu_\rho:\mathbb R^d\times\mathbb R^h\rightarrow \mathbb R$, parameterized by $\rho$, where $\nu_\rho(\phi_{x_i},\psi_{a_i})$ is a proto-value function whose Bellman residuals are estimates of the required stationary distribution ratios \cite{nachum2019dualdice}.  Given a trained $\nu_\rho$, the value of the RL-based DM's policy $\pi$ can be estimated by $J_{\text{DD}}(\pi_\lambda):=\sum_{i=1}^{|B|} r_i\cdot(\nu_{\rho^*}(\phi_{x_i},\psi_{a_i}) - \gamma\frac{\pi_\lambda(a'_i|x'_i)}{\pi_B(a'_i|x'_i)}\nu_{\rho^*}(\phi_{x'_i},\psi_{a'_i}))$.
Notice that this estimator assumes the knowledge of the behavior (data-generating) policy $\pi_B$ (which is the supervised DM in our setting). However, since the supervised DM is trained to optimize for a myopic reward, it can be overly deterministic. This can drive large fluctuations in propensity scores $\pi/\pi_B$, and high variance in $J_{\text{DD}}$. Instead, we use a \emph{behavior-agnostic} form of DualDICE which requires no estimation of $\pi_B$ (see \cite{nachum2019dualdice} for details).

The off-policy evaluation results generated by DualDICE are presented in Table~\ref{table:offline_eval} (second column). Since our use of DualDICE depends on language encoders $\phi$ and $\psi$, it cannot be used to evaluate the E2E model.\footnote{Extending DualDICE to evaluate E2E models requires a much more complex dual-function parameterization and will be left as future work.} Note that our off-policy value estimates of the RNN-based and the transformer-based models are generated using an RNN-based and a transformer-based behavior policy, respectively. 
% (which are mimicked by the RNN and Transformer models respectively). 
% According to DualDICE OPE, 
DualDice off-policy results show that SAQL-Reg-RNN and CAQL-RNN are among the best-performing RL-based policies (this coincides with on-policy evaluation results, see below). The offline performance of RNN-based models is consistently better than that of transformer-based models. This too is somewhat corroborated by on-policy evaluation, though this performance difference may be due in part to proto-function approximation error in DualDice caused by the inherent bias of the transformer-based data. 
% }

% \paragraph{\textbf{DM On-policy Evaluation}}{
\vspace{1mm}
\noindent
\textbf{DM On-policy Evaluation.}
% To assess the merit of our different models, 
We next conducted on-policy evaluation. Human evaluators were asked to conduct dialogues with our bot and rate the overall conversation experience on the same -3 to 7 scale used to collect training data. Evaluation was blind---evaluators did not know which model they were conversing with. Overall 200 dialogues for each model were rated. The results are presented in Table~\ref{table:offline_eval} (first column). SAQL-Reg-E2E and SAQL-Reg-RNN received the highest rating while SAQL-RNN performed worst. Notice that the models rated best by evaluators are trained with lower discount factors. We conjecture raters may be inherently biased to value myopic quality.%
\footnote{The RL models trained with lower discount factors behave closer to the supervised DM. Their DualDice estimates have lower variance and also tend to perform better.}  We also note the high-variance in rater evaluations across all models (a point discussed further below).
% }

\begin{table}
%\begin{small}
 \begin{tabular}{|l|l|c|c|} 
 \hline
 Model Type & Model  & On-policy  & Off-policy \\
 \hline\hline
 Supervised & RNN &   $5.75 \pm 1.99$ & $\mathbf{5.28\pm 0.44}$ \\ 
 \cline{2-4}
 & Transformer &  $5.67 \pm 2.18$ & $4.55\pm 0.37$ \\ 
 \hline
 \hline
 Stochastic & RNN &  $5.17 \pm 2.29$ & $5.13\pm 0.52$\\ 
  \cline{2-4}
 Actions & Reg-RNN &  $\mathbf{6.48 \pm 1.13}$ & $\mathbf{5.51\pm 0.39}$ \\ 
  \cline{2-4}
 & Transformer &  $5.71 \pm 2.26$ & $4.76\pm 0.40$ \\ 
  \cline{2-4}
 & Reg-Transformer &  $5.58 \pm 2.02$ & $4.73\pm 0.45$ \\ 
 \hline
 \hline
 Continuous & RNN &  $\mathbf{6.04 \pm 1.63}$ & $\mathbf{5.49\pm 0.41}$\\ 
  \cline{2-4}
 Actions & Transformer &   $5.86 \pm 1.79$ & $4.91\pm 0.38$  \\ 
  \cline{2-4}
 & Reg-Transformer &  $5.46 \pm 2.25$ & $4.78\pm 0.46$ \\ 
 \hline
 \hline
 E2E & Reg-Transformer &  $\mathbf{6.53} \pm 1.44$ & $--$\\
 \hline
\end{tabular}
\vspace{2mm}
%\end{small}
  \caption{Off-policy and on-policy raters evaluation results.}
  \label{table:offline_eval}
\vspace{-2mm}
\end{table}

\section{Live Experiment}
\label{sec:live_exp}

% In the previous section, we described the performance of our supervised and RL-based DM models as evaluated offline by dedicated human raters. 

Evaluation by human raters facilitates policy assessment in controlled settings and is necessary before deployment in a user-facing commercial product. However, dedicated human evaluators typically behave differently than real users. Specifically, the impact of conversation planning might be quite different with raters vs.\ real users. For example, raters might continue a conversation even after it reaches an awkward stage or they might not reflect a potential increase in user engagement after a successful focus change initiated by the bot. To gain an in-depth understanding of their impact on real users, we conducted a live experiment with our RL models. The Q-learning model that achieved the largest improvement in terms of user engagement was then fully deployed in the Google Assistant.
%\gal{We have something stronger than an experiment, a successful one that was launched. Are we allowed to say that?}

\subsection{Experimental Setup}
\label{ssec:online_setup}

To conduct a live experiment, we build on the \emph{dynamic composition} bot from \cite{szpektor2020dynamic} (Sec.~\ref{sec:dynamic_comp}). This bot is integrated with the Google Assistant, dubbed \emph{the assistant} below, and interacts with users in a real-time online setting. 

The experiment was conducted using an A/B testing protocol, in which a small percentage of assistant users were randomly sampled to interact with the bot using an RL-based DM while other users (same percentage) interact with the vanilla bot using a supervised DM. More precisely, the experiment was conducted with one control arm, with the transformer-based supervised model, and eight experiment arms with the architectures listed in Sec.~\ref{sec:offline}. %following RL architectures: SAQL-RNN, CAQL-RNN, SAQL-Transformer, CAQL-Transformer as well as the following CQL versions, SAQL-Reg-RNN, SAQL-Reg-Transformer, SAQL-Reg-Transformer and SAQL-Reg-E2E.
We use the supervised transformer model as a baseline as it was shown to outperform the supervised RNN in a previous live experiment.

Our experiment spanned the months of December 2021 and January 2022, during which user assignment to control/experiments remained constant. The experiment was transparent to the users, who could not distinguish between the different DMs. A conversation starts when a user triggers the experience by asking an animal related query (e.g., ``how does a lion sound?''). Once initiated, a conversation with a user could end if the bot predicted that its response is not of sufficient quality (i.e., the DM score is too low), if the user issued a query outside of the animal domain (\eg{} about the weather), or if the user issued a standard stop command. The last two options were handled by the assistant.

\subsection{Evaluation Metrics}

We measured daily user interaction with the assistant in the animal domain in both the experiment and control arms. To assess user engagement, we use several surrogate metrics that are directly measurable in the interaction logs.
We define a \emph{conversation} to be the succession of user and bot turns, starting with a triggering user turn.
% (Section~\ref{ssec:online_setup}).
The \emph{conversation length} is the number of turns (combined user and bot turns) in a conversation. We consider \emph{followup feedback} after each bot response, where followup refers to the next query, if any, after the bot response. Specifically, we distinguish:
\begin{itemize}
\item \emph{Cooperative responses} to bot questions, such as ``yes'' in response to a question proposing additional content (e.g., ``do you want to hear more?'') or ``Tell me about lions'' in response to a list selection question (e.g., ``which animal do you want to hear about next?'').
\item \emph{Non-cooperative responses} to bot questions, such as ``no'' in response to a question proposing additional content (e.g., ``do you want to learn about cheetahs?'').
% \item \emph{Neutral response} refers to changing the topic (answering “tell me about tigers” to the bot asking “do you want to hear more about lions?")
% \item \emph{End dialogue} refers to ending the current conversation (e.g., silence, “goodbye” or query not in the animal domain)
\item \emph{Explicit positive feedback}, which captures followup user queries with explicit gratitude, e.g., \ex{thank you} or \ex{wonderful}.
\item \emph{Explicit negative feedback}, reflecting followup user queries that contain negative feedback, such as \ex{stop} or \ex{shut up}.
\end{itemize}
For the last two metrics, we use predefined lists of positive and negative feedback phrases collected from user logs.

\begin{table*}
  %\todo{MK}{Debby, I reordered the metrics in this table from high to low significance (in my opinion). WDYT? - I would leave conversation length as it is different from the other feedback related metrics (and even add an hline below. In terms of importance, I'd say that pos/neg response are more important since the explicit feedback such as thank you and shut up are a lot more rare.}
  %\gal{I actually don't think we want that many columns in the main body of the paper (OK to add in the appendix). Instead, we can say at the beginning that we try these variants and limit to interesting differences. For example, if we drop all CQL columns for the table, do we really lose much? We can just describe what happens in this case in a single sentence}
  \caption{Mean relative change of experiment vs. the control metrics. Here, T stands for transformer; green changes are desirable, red changes less so (to varying degrees).}.
  \label{tab:main_results}
  %\small
    \begin{tabular}{l|c|c|c|c|c}
        Metric & SAQL-RNN & SAQL-T &  CAQL-RNN  & CAQL-T & SAQL-Reg-E2E \\
        \hline
        Conversation length & \PosDiff{+30\%} & \PosDiff{+23\%} &  \PosDiff{+14\%} & \PosDiff{+18\%} &   \NegDiff{-0.7\%}\\
        \hline
        Cooperative response & \PosDiff{+8\%} &  \NegDiff{-6.8\%} &  \NegDiff{-5.8\%} & \NegDiff{-4\%} &  \NegDiff{-8\%}  \\
        Non-cooperative response & \NegDiff{+112\%} &  \NegDiff{+178\%} &  \NegDiff{+54\%} & \NegDiff{+120\%} &  \NegDiff{+41\%}  \\
        Explicit positive feedback & \PosDiff{+32\%} &  \PosDiff{+9.7\%} & \NegDiff{-20\%} & \PosDiff{+6.8\%} & \NegDiff{-6\%}  \\
        Explicit negative feedback & \PosDiff{-18\%} &  \NegDiff{+8.6\%} & \NegDiff{+1\%} & \PosDiff{-14\%} &  \NegDiff{+27\%}  \\
        % Conversation end & \PosDiff{-27\%} & \PosDiff{-3\%} & & & \PosDiff{-11\%} & & & \NegDiff{+3\%}\\
        % Neutral response & +18\% & +81\% & & & +46\% & & & +19\%  \\
    \end{tabular}
%    \vspace{-5mm}
\end{table*}

\subsection{Main Results}
The average relative change in metrics across all experiments w.r.t.\ the control is shown in Table~\ref{tab:main_results} (for CQL variants, see Table~\ref{tab:main_results_full} in the appendix). Interestingly, these results differ from the rater online evaluations in Table~\ref{table:offline_eval}, demonstrating the substantial distinction between the behaviors of raters and real users.
% \footnote{This discrepancy may be, in part, due to the high variance in rater evaluations.}
This discrepancy may be, in part, due to the high variance in rater evaluations.
Surprisingly, SAQL-Reg-E2E performs worst and is slightly outperformed by the supervised baseline. The E2E model behaves quite conservatively, similar to a supervised model, avoiding pivoting to other animals and changing the type of offered content (e.g., sounds, facts, quizzes). 
Such conservative behavior may be caused by the lower discount factor $\gamma=0.8$ used, making its expected trajectory horizon shorter. %Its expected trajectory horizon becomes shorter (5 steps) than the other RL models (20 steps; mostly, the discount factor is 0.95), therefore, its behaviour will be more myopic and closer to the supervised model.
This might be preferred by raters who tend to evaluate bot responses more myopically; but at the same time, it provides a more boring, less engaging experience for users. 
%\gal{Do we have a conjecture for this? E.g., because raters are different, doing the absolute best on that can be detrimental
The different SAQL models outperform their CAQL counterparts. A similar conclusion was drawn in \cite{zhao-etal-2019-rethinking}, where discrete latent actions were deemed to be more suitable than continuous actions for dialogue agents.

Overall, we find that SAQL-RNN performs best w.r.t.\ our main metrics, conducting longer, more engaging conversations. It increases conversation length by $32\%$, while also increasing user engagement as captured by multiple metrics. We see an increase of $8\%$ in cooperative responses to bot questions. While there is also a large increase of non-cooperative responses ($112\%$), this is expected as the SAQL-RNN agent takes more risks by asking pivoting questions, generating many more occasions for non-cooperative user reactions. While the user may not be interested in the conversational direction proposed by the bot (e.g., pivoting to another animal), the user often continues engaging in a dialogue about animals. For example, 
% in the conversation 
in Fig.~\ref{fig:exp_dialogue}, the user provides a non-cooperative answer in the 3rd turn. As a result, the bot modifies its plan and asks the user to choose the next conversation focus, 
to which the user responds positively.
% which the user cooperates with.
%\todo{Debby}{Consider explaining why this is ok and that future work should fix that} \gal{We {\bf have to} explain this}.
In addition, some followup user queries contain explicit positive or negative feedback. While an order of magnitude fewer than other followups, they offer a \emph{direct} measure of user (dis)satisfaction. SAQL-RNN increases explicit positive feedback by $32\%$ and reduces negative feedback by $18\%$.

Our CQL variants of the different models indeed behave more ``conservatively,'' closer to supervised model behavior. This translates into smaller changes in conversation length and user feedback metrics (see appendix). Interestingly, using the transformer (vs.\ RNN) state encoding does not improve SAQL performance, unlike in the supervised setting, where transformer-based candidate selection is superior: the RNN state representation seems sufficient for RL. For this reason, we focus 
our analysis 
on SAQL-RNN below.

\subsection{Qualitative Analysis of the RL DM}
%\gal{You want RL in the ttile of this sub-section. Maybe 'Qualitative Characteristics of the RL Planner' or something like that}
To improve user engagement while conducting longer conversations,  SAQL-RNN uses several planning strategies. First, it ends $20\%$ more turns in questions relative to the control, prompting the user to choose additional content (e.g., learn more animal facts, hear another animal sound). While we observe an increase in cooperative responses, the \emph{cooperation rate} to bot questions drops by $9.5\%$. Although this may seem problematic, this is actually a result of a favorable policy learned by our bot: by taking more risks in eliciting a user's preference for the next steps, SAQL-RNN achieves an overall improved user experience, as measured via increased conversation length, combined with a noticeable increase in explicit positive feedback and a decrease in negative feedback. 

A second planning strategy is to better exploit content diversity, including facts, sounds, quizzes, yes/no questions, open questions, etc. On average, SAQL-RNN uses $26\%$ more unique providers per conversation than the supervised transformer-based model.

Two additional planning strategies are related to the existence of two sub-dialogues with different characteristics. Dialogues around animal sounds are poorer in content and exhibit entity pivoting at every turn (after playing the sound of a given animal, we can either suggest the sound of a different animal or quiz the user about other animal sounds). In contrast, dialogues around animal facts typically contain richer content and a greater conversation depth. We observe that SAQL-RNN favors the richer experience of the latter, selecting $31\%$ more fact-related content.

Lastly, we observe that the average conversation breadth of dialogues conducted by SAQL-RNN is lower (it generates $13\%$ fewer focus-pivoting turns). This is a consequence of fact dialogues having less breadth. However, when restricting analysis to fact dialogues, SAQL-RNN exhibits $60\%$ more focus-pivoting turns.

%\gal{You can't really expect the user to contrast this across pages. Put them one next to each other and build it as you did for the announcement}
Some of these strategies are exemplified by the sample conversation in Fig.~\ref{fig:exp_dialogue}, generated by the SAQL-RNN model, which we contrast with Fig.\ref{fig:ctl_dialogue}, conducted by the supervised transformer. Both conversations start with the same 2 turns. In the 3rd turn, after a non-cooperative user response, the transformer pivots back to sounds to maximize ``immediate'' user interest. By contrast, the RL model tries to pivot to facts for a richer conversational experience, suggesting that the user choose the next animal. We also observe that the RL conversation includes more types of content, such as sounds, facts, quizzes, yes/no and open questions.

\section{Conclusion}
\label{sec:conclusions}

In this work we tackled the formidable task of building a rich, open-ended conversational bot that is deployed in the  challenging setting of a real-time, global commercial assistant. Our approach relies on the framework of reinforcement learning, using a novel state representation based on the succinct embedding of a supervised language model and an RL algorithm that allows for a dynamic action space at each stage of the conversation.
Ours is one of the few examples of RL-based conversational systems deployed in the wild at scale, and the substantial advantages demonstrated over the SOTA supervised model validates the decades-long premise that the dynamic planning ability of RL is a natural fit for the design of rich dialogue agents. 

An interesting insight from our live experiment highlights the power of RL to take counter-intuitive actions: an increase in non-cooperative responses, a seemingly negative phenomenon, is simply a tool with which the agent may elicit a user's preference for the next phase of the conversation. This leads to a positive conversational experience on average, with a measurable increase both in conversation length and positive feedback. We hope to discover other dialogue strategies that drive ``great'' conversations as we shift to learning models directly from rich user signals.

\bibliographystyle{ACM-Reference-Format}
\bibliography{ref}

\newpage

\appendix
\section{Appendix}
\subsection{DM Training Hyper-Parameters}
Our supervised and RL models were trained with the following hyper-parameters.

The supervised RNN model is trained with a learning rate of 0.0001, batch size of 16, a dropout probability of 0.2 and $200$K training steps. Its architecture includes a GRU layer with 200 units.

For the supervised transformer model, we use the \texttt{BERT-Medium} checkpoint\footnote{\url{https://github.com/google-research/bert}} having uncased vocabulary, hidden dimension $H=512$, $L=8$ transformer layers, and $A=8$ attention heads per layer. This model was trained for $20000$ steps with a global batch size of $768$ divided among $8$ TPUv3 chips, using the Adam optimizer with an initial learning rate of $\epsilon=5\cdot 10^{-5}$ (decayed to zero), $\beta_1=0.9$, $\beta_2=0.999$, and $\epsilon=10^{-6}$. 

For the RL models, SAQL-RNN, CAQL-RNN, SAQL-Reg-RNN, and CAQL-Reg-RNN use a fully connected feedforward network for $Q$-function approximation. These networks are composed of $3$ layers and each layer is composed of $1024$ RELU units. The SAQL-RNN network is trained using a learning rate $\epsilon=7\cdot10^{-5}$ and $k=2M$ steps. Warmstarted with the SAQL-RNN weights, the SAQL-Reg-RNN network is trained with $\epsilon=5\cdot10^{-6}$ and $k=2.4M$. The CAQL-RNN network is trained with $\epsilon=5\cdot10^{-5}$ and $k=2M$, where the inner maximization problem uses a GA learning rate of $\epsilon_{\text{GA}}=1\cdot10^{-6}$ and runs for a maximum of $k_{\text{GA}}=25$ steps. Warmstarted with the CAQL-RNN weights, the CAQL-Reg-RNN network is trained with $\epsilon=3\cdot10^{-6}$ and $k=2M$. SAQL-Transformer, SAQL-Reg-Transformer, CAQL-Transformer, and CAQL-Reg-Transformer all follow almost the same settings as SAQL-RNN, SAQL-Reg-RNN, CAQL-RNN, and CAQL-Reg-RNN but are trained with $\epsilon=3\cdot10^{-4}$ and $k=4M$, $\epsilon=1\cdot10^{-5}$ and $k=2.5M$, $\epsilon=3\cdot10^{-5}$ and $k=3M$, $\epsilon=1\cdot10^{-5}$ and $k=3M$ respectively. The models are trained with batch size $|B|=32$, and in all two-step CQL regularized models the regularization coefficient $\alpha$ is $0.1$.\footnote{Compared with learning rates, tuning $\alpha$ does not have much effect on model training.} SAQL-Reg-E2E was trained with $k=320,000$, $\epsilon=5\cdot10^{-5}$, $|B|=48$, $\gamma=0.8$ and $\alpha=0.01$. All these hyper-parameters are chosen from the best settings of their corresponding grid-search optimization.

\begin{table*}[th!]
  \caption{Mean relative change of experiments vs. the control metrics. Here, T stands for transformer.}
  \label{tab:main_results_full}
  \small
    \begin{tabular}{l|c|c|c|c|c|c|c|c}
        Metric & SAQL-RNN & SAQL-Reg-RNN & SAQL-T & SAQL-Reg-T & CAQL-RNN  & CAQL-T & CAQL-Reg-T & SAQL-Reg-E2E \\
        \hline
        Conversation length & \PosDiff{+30\%} & \PosDiff{+3.6\%} & \PosDiff{+23\%} & \PosDiff{+19\%} & \PosDiff{+14\%} & \PosDiff{+18\%} & \PosDiff{+8.9\%} &  \NegDiff{-0.7\%}\\
        \hline
        Cooperative response & \PosDiff{+8\%} & \NegDiff{-20\%} & \NegDiff{-6.8\%} & \NegDiff{-0.3\%} & \NegDiff{-5.8\%} & \NegDiff{-4\%} & \NegDiff{-4.3\%} & \NegDiff{-8\%}  \\
        Non-cooperative response & \NegDiff{+112\%} & \NegDiff{+75\%} & \NegDiff{+178\%} & \NegDiff{+125\%} & \NegDiff{+54\%} & \NegDiff{+120\%} & \NegDiff{+130\%} & \NegDiff{+41\%}  \\
        Explicit positive feedback & \PosDiff{+32\%} & \PosDiff{+42\%} & \PosDiff{+9.7\%} & \PosDiff{+0.4\%} & \NegDiff{-20\%} & \PosDiff{+6.8\%} & \PosDiff{+5.8\%} & \NegDiff{-6\%}  \\
        Explicit negative feedback & \PosDiff{-18\%} & \PosDiff{-7\%} & \NegDiff{+8.6\%} & \PosDiff{-7.7\%} & \NegDiff{+1\%} & \PosDiff{-14\%} & \NegDiff{+15\%}  & \NegDiff{+27\%}  \\
        
        % Conversation end & \PosDiff{-27\%} & \PosDiff{-3\%} & & & \PosDiff{-11\%} & & & \NegDiff{+3\%}\\
        
        % Neutral response & +18\% & +81\% & & & +46\% & & & +19\%  \\
    \end{tabular}
\end{table*}

\subsection{Full Live Experiment Results}

The average relative change in the live experiment metrics of the experiments w.r.t the control is shown in Table~\ref{tab:main_results_full} for all models, including the CQL variants.

%\begin{figure*}
%  \includegraphics[width=2\columnwidth]{figs/xtalk-rater-ui}
%\caption{User interface that human raters use for scoring the candidate utterances.}
%\todo{MK}{Debby, feel free to move this figure to Appendix or remove if not allowed for publication. - Debby: I don't think it will be allowed. Also, I think we should reduce the content related to section 3, which was already covered by the WWW publication.}
%  \label{fig:xtalk_rater_ui}
%\end{figure*}

\end{document}